\documentclass[11pt,letterpaper]{article}
\usepackage{naaclhlt2016}
\usepackage{times}
\usepackage{latexsym}
\naaclfinalcopy 

\usepackage{graphicx}
\usepackage{color}
\usepackage{booktabs}
\usepackage{subcaption}
\usepackage{multirow}
\usepackage{amsmath}
\usepackage{amssymb}

\usepackage{pgfplots}
\usepackage{tikz}
\usepackage{tkz-graph}
\usepackage{algorithmic}

\usepackage{graphicx}
\usepackage[export]{adjustbox}

\newif\ifdraft
\drafttrue

\title{SemEval-2016 Task 3: Community Question Answering}

\author{{Preslav Nakov}\hspace*{2.5mm}
        {Llu\'{i}s M\`arquez}\hspace*{2.5mm}
        {Alessandro Moschitti}\smallskip\\
        {\bf Walid Magdy}\hspace*{2mm}
        {\bf Hamdy Mubarak}\hspace*{2mm}
        {\bf Abed Alhakim Freihat}
        \smallskip\\
        ALT Research Group, Qatar Computing Research Institute, HBKU\\\smallskip\\
        {\bf James Glass}\\
        MIT Computer Science and Artificial Intelligence Laboratory\\\smallskip\\
        {\bf Bilal Randeree}\\
        Qatar Living\\}

\date{\today}

\begin{document}
\maketitle

\begin{abstract}
This paper describes 
the SemEval--2016 Task 3 on Community Question Answering, which we offered in English and Arabic.
For English, we had three subtasks: \emph{Question--Comment Similarity} (subtask A), \emph{Question--Question Similarity} (B), and \emph{Question--External Comment Similarity} (C). For Arabic, we had another subtask: \emph{Rerank the correct answers for a new question} (D). 
Eighteen teams participated in the task, submitting a total of 95 runs (38 primary and 57 contrastive) for the four subtasks. A variety of approaches and features were used by the participating systems to address the different subtasks, which are summarized in this paper. The best systems achieved an official score (MAP) of 79.19, 76.70, 55.41, and 45.83 in subtasks A, B, C, and D, respectively. 
These scores are significantly better than those for the baselines that we provided.
For subtask A, the best system improved over the 2015 winner by 3 points absolute in terms of Accuracy.
\end{abstract}

\section{Introduction}
\label{sec:intro}

Building on the success of SemEval--2015 Task 3 ``Answer Selection in Community Question Answering''\footnote{http://alt.qcri.org/semeval2015/task3} \cite{nakov-EtAl:2015:SemEval}, we run an extension in 2016, which covers a full task on \emph{Community Question Answering} (CQA) and which is, therefore, closer to the real application needs. All the information related to the task, data, participants, results and publications can be found on the SemEval--2016 Task 3 website.\footnote{http://alt.qcri.org/semeval2016/task3}

CQA forums such as Stack Overflow\footnote{http://stackoverflow.com/} and Qatar Living\footnote{http://www.qatarliving.com/forum}, are gaining popularity online. These forums are seldom moderated, quite open, and thus they typically have little restrictions, if any, on who can post and who can answer a question. On the positive side, this means that one can freely ask any question and can then expect some good, honest answers. On the negative side, it takes effort to go through all possible answers and to make sense of them. For example, it is not unusual for a question to have hundreds of answers, which makes it very time-consuming for the user to inspect and to winnow through them all. The present task could help to automate the process of finding good answers to new questions in a community-created discussion forum, e.g., by retrieving similar questions in the forum and by identifying the posts in the comment threads of those similar questions that answer the original question well.

In essence, the main CQA task can be defined as follows: ``given (\emph{i}) a new question and (\emph{ii}) a large collection of question-comment threads created by a user community, rank  the comments that are most useful for answering the new question''.

The test question is new with respect to the collection, but it is expected to be related to one or several questions in the collection. The best answers can come from different question-comment threads. In the collection, the threads are independent of each other and the lists of comments are chronologically sorted and contain some meta information, e.g., date, user, topic, etc.

The comments in a particular thread are intended to answer the question initiating that thread, but since this is a resource created by a community of casual users, there is a lot of noise and irrelevant material, apart from informal language usage and lots of typos and grammatical mistakes. Interestingly, the questions in the collection can be semantically related to each other, although not explicitly.

Our intention was not to run just another regular Question Answering task. Similarly to the 2015 edition, we had three objectives: (\emph{i})~to focus on semantic-based solutions beyond simple ``bag-of-words'' representations and ``word matching'' techniques; (\emph{ii})~to study the new natural language processing (NLP) phenomena arising in the community question answering scenario, e.g., relations between the comments in a thread, relations between different threads and question-to-question similarity; and (\emph{iii})~ to facilitate the participation of non IR/QA experts to our challenge. The third point was achieved by explicitly providing the set of potential answers---the search engine step was carried out by us---to be (re)ranked and by defining two optional subtasks apart from the main CQA task. Subtask A (\emph{Question-Comment Similarity}): given a question from a question-comment thread, rank the comments according to their relevance (similarity) with respect to the question; Subtask B (\emph{Question-Question Similarity}): given the new question, rerank all similar questions retrieved by a search engine, assuming that the answers to the similar questions should be answering the new question too.

Subtasks A and B should give participants enough tools to create a CQA system to solve the main task. Nonetheless, one can approach CQA without necessarily solving the two tasks above. Participants were free to use whatever approach they wanted, and the participation in the main task and/or the two subtasks was optional. A more precise definition of all subtasks can be found in Section~\ref{sec:task}.

Keeping the multilinguality from 2015, we provided data for two languages: English and Arabic. For English, we used real data from the community-created Qatar Living forum. The Arabic data was collected from medical forums, with a slightly different procedure. 
We only proposed the main ranking CQA task on this data, i.e., finding good answers for a given new question. 

Finally, we provided training data for all languages and subtasks with human supervision. All examples were manually labeled by a community of annotators in a crowdsourcing platform. The datasets and the annotation procedure are described in Section~\ref{sec:datasets}, and some examples can be found in Figures~\ref{EnglishXMLFig}~and~\ref{fig:arabicsample}.

The rest of the paper is organized as follows:
Section~\ref{sec:related} introduces some related work.
Section~\ref{sec:task} gives a more detailed definition of the task.
Section~\ref{sec:datasets} describes the datasets and the process of their creation.
Section~\ref{sec:scoring} explains the evaluation measures.
Section~\ref{sec:results} presents the results for all subtasks and for all participating systems.
Section~\ref{sec:features} summarizes the main approaches and features used by these systems. 
Finally, Section~\ref{sec:conclusion} offers some further discussion
and presents the main conclusions.

\section{Related Work}
\label{sec:related}

Our task goes in the direction of passage reranking,
where automatic classifiers are normally applied to pairs of questions and answer passages to derive a relative order between passages, e.g., see \cite{abs-cs-0605035,Jeon:2005:FSQ:1099554.1099572,shen-lapata:2007:EMNLP-CoNLL2007,Moschitti_et_al:ACL2007,severyn2015sigir,CIKM2008,Tymoshenko:2015:AIS:2806416.2806490,Tymoshenko:NAACL:2016,Surdeanu2008}.
In recent years, many advanced models have been developed for automating answer selection, producing a large body of work.
For instance,  \newcite{wang:2007} proposed a probabilistic quasi-synchronous grammar to learn syntactic transformations from the question to the candidate answers; \newcite{heilman:naacl:2010} used an algorithm based on Tree Edit Distance (TED) to learn tree transformations in pairs; \newcite{wang_manning:acl:2010} developed a probabilistic model to learn tree-edit operations on dependency parse trees; and \newcite{yao:naacl:2013} applied linear chain CRFs with features derived from TED to automatically learn associations between questions and candidate answers.
One interesting aspect of the above research is the need for syntactic structures;
this is also corroborated in
\cite{sigir12,severyn-moschitti:2013:EMNLP}.  Note that answer selection can use models for textual entailment, semantic similarity, and for natural language inference in general. 

Using information about the thread is another important direction.
In the 2015 edition of the task, the top participating systems used thread-level features, in addition to the usual local features that only look at the question--answer pair.
For example, the second-best team, HITSZ-ICRC, used as a feature the position of the comment in the thread, whether the answer is first, whether the answer is last
\cite{hou-EtAl:2015:SemEval1}.
Similarly, the third-best team, QCRI, used features 
that model a comment in the context of the entire comment thread, focusing on user interaction~\cite{nicosia-EtAl:2015:SemEval}.
Finally, the fifth-best team, ICRC-HIT, treated the answer selection task as a sequence labeling problem and proposed recurrent convolution neural networks to recognize good comments \cite{zhou-EtAl:2015:SemEval}.

In a follow-up work, \newcite{zhou-EtAl:2015:ACL-IJCNLP} included long-short term memory (LSTM) units in their convolutional neural network to learn the classification sequence for the thread. In parallel, \newcite{barroncedeno-EtAl:2015:ACL-IJCNLP} exploited the dependencies between the thread comments to tackle the same task. This was done by designing features that look globally at the thread and by applying structured prediction models, such as Conditional Random Fields \cite{Lafferty01}. 

This research direction was further extended by \newcite{joty:2015:EMNLP}, who used the output structure at the thread level in order to make more consistent global decisions. For this purpose, they modeled the relations between pairs of comments at any distance in the thread, and they combined the predictions of local classifiers in a graph-cut and in an ILP frameworks. 

Finally, \newcite{Joty:2016:NAACL} proposed two novel joint learning models that are on-line and integrate inference within the learning process. The first one jointly learns two node- and edge-level MaxEnt classifiers with stochastic gradient descent and integrates the inference step with loopy belief propagation. The second model is an instance of fully connected pairwise CRFs (FCCRF). The FCCRF model significantly outperforms all other approaches and yields the best results on the task (SemEval-2015 Task 3) to date. Crucial elements for its success are the global normalization and an Ising-like edge potential.

\section{Definition of the Subtasks}
\label{sec:task}

The challenge was structured as a set of four different and independent subtasks. Three of them (A, B and C) were offered for English, while the fourth one (D) was offered for Arabic. We describe them below in detail. 
In order to make the subtask definitions more clear, we also provide some high-level information about the datasets we used (they will be described in more detail later in Section~\ref{sec:datasets}).

The English data comes from the Qatar Living forum, which is organized as a set of seemingly independent question--comment threads.  In short, for subtask A we annotated the comments in a question-thread as ``Good'', ``PotentiallyUseful'' or ``Bad'' with respect to the question that started the thread. Additionally, given original questions we retrieved related question--comment threads and we annotated the related questions as ``PerfectMatch'', ``Relevant'', or ``Irrelevant'' with respect to the original question (subtask B). We then annotated the comments in the threads of related questions as ``Good'', ``PotentiallyUseful'' or ``Bad'' with respect to the original question (subtask C).

For Arabic, the data was extracted from medical forums and has a different format. Given an original question, we retrieved pairs of the form (related\_question, answer\_to\_the\_related\_question). These pairs were annotated as ``Direct'' answer, ``Relevant'' and ``Irrelevant'' with respect to the original question.

\paragraph{English subtask A} \emph{Question-Comment Similarity}.
Given a question $Q$ and its first ten comments\footnote{We limit the number of comments we consider to the first ten only in order to spare some annotation efforts.} in the question thread ($c_1,\dots,c_{10}$), the goal is to rank these ten comments according to their relevance with respect to the question.

Note that this is a ranking task, not a classification task; we use mean average precision (MAP) as an official evaluation measure. This setting was adopted as it is closer to the application scenario than pure comment classification. For a perfect ranking, a system has to place all ``Good'' comments above the ``PotentiallyUseful'' and ``Bad'' comments; the latter two are not actually distinguished and are considered ``Bad'' in terms of evaluation. 

Note also that subtask A this year is the same as subtask A at SemEval-2015 Task 3, but with slightly different annotation and evaluation measure.

\paragraph{English subtask B} \emph{Question-Question Similarity}.
Given a new question $Q$ (aka original question) and the set of the first ten related questions from the forum ($Q_1,\dots,Q_{10}$) retrieved by a search engine, the goal is to rank the related questions according to their similarity with respect to the original question. 

In this case, we consider the ``PerfectMatch'' and ``Relevant'' questions both as good (i.e., we do not distinguish between them and we will consider them both ``Relevant''), and they should be ranked above the ``Irrelevant'' questions. 
As in subtask A, we use MAP as the official evaluation measure. To produce the ranking of related questions, participants have access to the corresponding related question-thread.\footnote{Note that the search engine indexes entire Web pages, and thus, the search engine has compared the original question to the related questions together with their comment threads.} 
Thus, being more precise, this subtask could have been named \emph{Question --- Question+Thread Similarity}.

\paragraph{English subtask C} \emph{Question-External Comment Similarity}.
Given a new question $Q$ (aka the original question), and the set of the first ten related questions ($Q_1,\dots,Q_{10}$) from the forum retrieved by a search engine, each associated with its first ten comments appearing in its thread ($c_1^1,\dots,c_1^{10},\dots,c_{10}^1,\dots,c_{10}^{10}$), the goal is to rank the 100 comments $\{c_i^j\}_{i,j=1}^{10}$ according to their relevance with respect to the original question $Q$. 

This is the main English subtask.
As in subtask A, we want the ``Good'' comments to be ranked above the ``PotentiallyUseful'' and ``Bad'' comments, which will be considered just bad in terms of evaluation. Although, the systems are supposed to work on 100 comments, we take an application-oriented view in the evaluation,
assuming that
users would like to have good comments concentrated in the first ten positions.
We believe users care much less about what happens in lower positions (e.g., after the 10th) in the rank, as they typically do not ask for the next page of results in a search engine such as Google or Bing. This is reflected in our primary evaluation score, MAP, which we restrict to consider only the top ten results in subtask C.

\paragraph{Arabic subtask D} \emph{Rank the correct answers for a new question}.
Given a new question $Q$ (aka the original question), the set of the first 30 related questions retrieved by a search engine, each associated with one correct answer ($(Q_1,c_1)\dots,(Q_{30},c_{30})$), the goal is to rank the 30 question-answer pairs according to their relevance with respect to the original question. We want the ``Direct'' and the ``Relevant'' answers to be ranked above the ``Irrelevant'' answers; the former two are considered ``Relevant'' in terms of evaluation. We evaluate the position of ``Relevant'' answers in the rank, therefore, this is again a ranking task.

Unlike the English subtasks, here we use 30 answers since the retrieval task is much more difficult, leading to low recall, and the number of correct answers is much lower. Again, systems were evaluated using MAP, restricted to the top-10 results.

\section{Datasets}
\label{sec:datasets}



As we mentioned above, the task is offered for two languages, English and Arabic.
Below we describe the data for each language.


\subsection{English Dataset}
\label{sec:englishdataset}

We refer to the English data as the CQA-QL corpus;
it is based on data from the Qatar Living forum.

The English data is organized with focus on the main task, which is subtask C, but it contains annotations for all three subtasks.
It consists of a list of original questions,
where for each original question
there are ten related questions from Qatar Living,
together with the first ten comments from their threads.
The data is annotated with the relevance of each related question with respect to the original question (subtask B),
as well as with the relevance of each comment with respect to the related (subtask A) and also with respect to the original question (subtask C).

To build the dataset, we first selected a set of questions to serve as original questions. In a real-world scenario those would be questions that were never asked before; however, here we used existing questions 
from Qatar Living. For the training and for the development datasets, we used questions from SemEval-2015 Task 3 \cite{nakov-EtAl:2015:SemEval}, while we used new Qatar Living questions for testing.

From each original question, we generated a query, using the question's subject (after some word removal if the subject was too long). Then, we executed the query in Google, limiting the search to the Qatar Living forum, and we collected up to 200 resulting question-comment threads as related questions.
Afterwards, we filtered out threads with less than ten comments as well as those for which the question was more than 2,000 characters long. Finally, we kept the top-10 surviving threads, keeping just the first 10 comments in each thread.

We formatted the results in XML with UTF-8 encoding, adding metadata for the related questions and for their comments; however, we did not provide any meta information about the original question, in order to emulate a scenario where it is a new question, never asked before in the forum. In order to have a valid XML, we had to do some cleansing and normalization of the data. We added an XML format definition at the beginning of the XML file and made sure it validates.

We provided a split of the data into three datasets: training, development, and testing.
A dataset file is a sequence of original questions (OrgQuestion), where each question has a subject, a body (text), and a unique question identifier (ORGQ\_ID). Each such original question is followed by ten threads, where each thread has a related question (according to the search engine results) and its first ten comments.

Each related question (RelQuestion) has a subject and a body (text), as well as the following attributes:
\begin{itemize}
\itemsep0em 
\item RELQ\_ID: question identifier;
\item RELQ\_RANKING\_ORDER: the rank of the related question in the list of results returned by the search engine for the original question;\footnote{This is the rank of the thread in the original list of Google results, before the thread filtering; see above.}
\item RELQ\_CATEGORY: the question category, according to the Qatar Living taxonomy;\footnote{Here are some examples of Qatar Living categories: Advice and Help, Beauty and Style, Cars and driving, Computers and Internet, Doha Shopping, Education, Environment, Family Life in Qatar, Funnies, Health and Fitness, Investment and Finance, Language, Moving to Qatar, Opportunities, Pets and Animals, Politics, Qatar Living Lounge, Qatari Culture, Salary and Allowances, Sightseeing and Tourist attractions, Socialising, Sports in Qatar, Visas and Permits, Welcome to Qatar, Working in Qatar.}
\item RELQ\_DATE: date of posting;
\item RELQ\_USERID: identifier of the user asking the question;
\item RELQ\_USERNAME: name of the user asking the question;
\item RELQ\_RELEVANCE2ORGQ: human assessement on the relevance this RelQuestion thread with respect to OrgQuestion. This label can take one of the following values:
	\begin{itemize}
	\itemsep0em     
 		\item \emph{PerfectMatch}: RelQuestion matches OrgQuestion (almost) perfectly; at test time, this label is to be merged with Relevant;
  		\item \emph{Relevant}: RelQuestion covers some aspects of OrgQuestion;
  		\item \emph{Irrelevant}: RelQuestion covers no aspects of OrgQuestion.
  	\end{itemize}
\end{itemize}

\begin{figure*}[ht]
\centering
\includegraphics [scale=0.55]{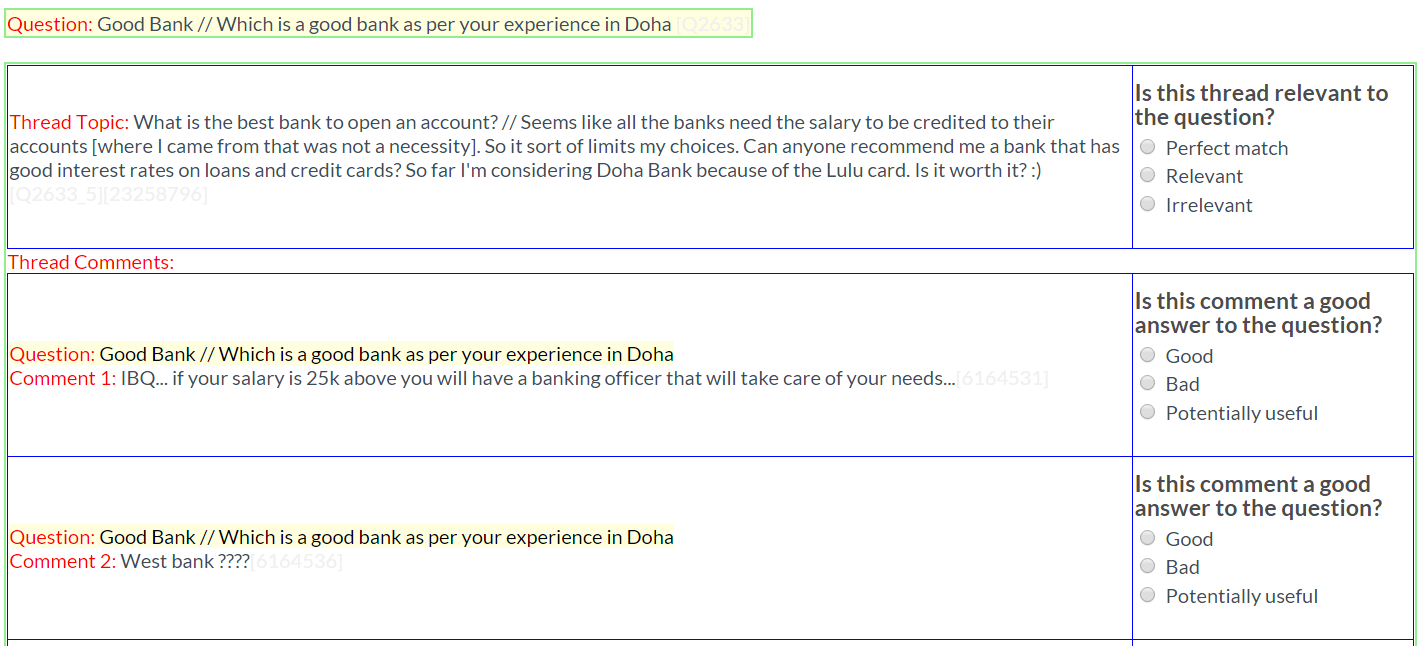}
\caption{Screenshot for the first English annotation job, collecting labels for subtasks B and C.}
\label{CFAnnotationFig}
\end{figure*}

\begin{figure*}[ht]
\centering
\includegraphics [width=\textwidth]{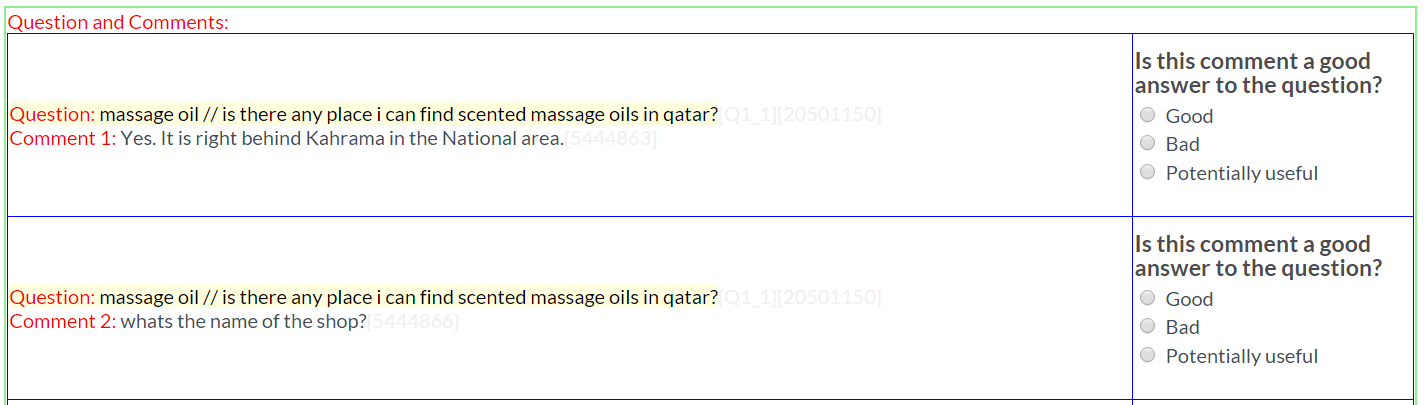}
\caption{Screenshot of the second English annotation job, collecting labels for subtask A.}
\label{CFAnnotation2Fig}
\end{figure*}

\begin{figure*}[ht]
\centering
\includegraphics [width=\textwidth, frame]{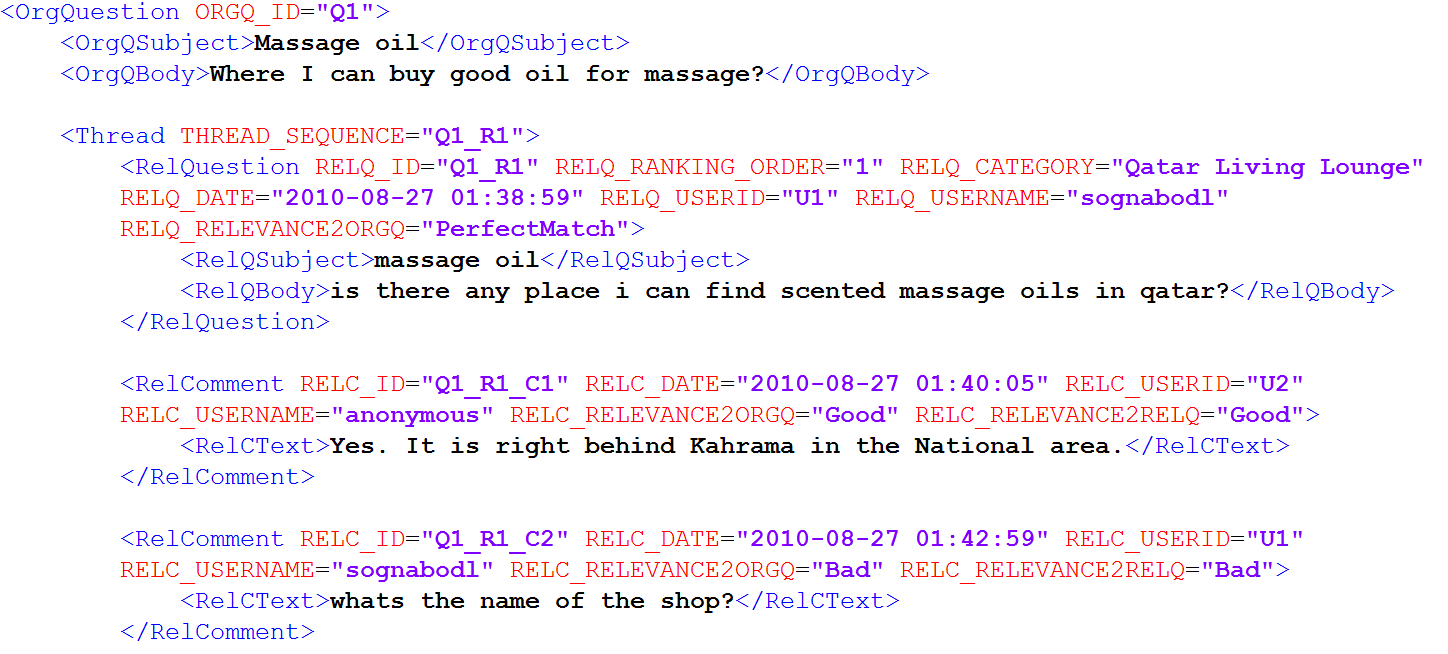}
\caption{Annotated English question from the CQA-QL corpus. Shown are the first two comments only.}
\label{EnglishXMLFig}
\end{figure*}

\begin{table*}[ht]
\small
\begin{center}
\begin{tabular}{l@{}rrrrrr}
\bf Category & \bf Train & \bf Train & \bf Train+Dev+Test & \bf Dev & \bf Test & \bf Total\\
  & \bf (1st part) & \bf (2nd part) & \bf (from SemEval 2015) & & & \\
\hline
\hline
\bf Original Questions & \bf 200 & \bf 67 & \multicolumn{1}{c}{\bf --} & \bf 50 & \bf 70 & \bf 387 \\
\bf Related Questions & \bf 1,999 & \bf 670 & \bf 2,480+291+319 & \bf 500 & \bf 700  & \bf 6,959 \\
\it -- Perfect Match & \it 181 & \it 54 & \multicolumn{1}{c}{--} & \it 59 & \it 81 &  \it 375 \\
\it -- Relevant & \it 606 & \it 242 & \multicolumn{1}{c}{--} & \it 155 & \it 152 &  \it 1,155 \\
\it -- Irrelevant & \it 1,212 & \it 374 & \multicolumn{1}{c}{--} & \it 286 & \it 467 &  \it 2,339 \\
&  &  & \bf  & \bf  & \bf \\
\bf Related Comments & \bf 19,990 & \bf 6,700 & \multicolumn{1}{c}{\bf --} & \bf 5,000  & \bf 7,000  & \bf 38,690\\
\bf (with respect to Original Question) &   &  & &   &   &  \\
\it -- Good & \it 1,988 & \it 849 & \multicolumn{1}{c}{--} & \it 345 &  \it 654 &  \it 3,836\\
\it -- Bad & \it 16,319 & \it 5,154 & \multicolumn{1}{c}{--} & \it 4,061 &  \it 5,943  &  \it 31,477 \\
\it -- Potentially Useful & \it 1,683 &  \it 697 & \multicolumn{1}{c}{--} & \it 594  &  \it 403 &  \it 3,377 \\
&  &  & \bf  & \bf  & \bf \\
\bf Related Comments & \bf 14,110 & \bf 3,790 & \bf 14,893+1,529+1,876 & \bf 2,440 & \bf 3,270 & \bf 41,908 \\
\bf (with respect to Related Question) &   &  &   & &   &  \\
\it -- Good & \it 5,287 & \it 1,364 & \it 7,418+813+946 & \it 818 &  \it 1,329 &  \it 17,975 \\
\it -- Bad & \it 6,362 & \it 1,777 & \it 5,971+544+774 & \it 1,209 &  \it 1,485 &  \it 18,122 \\
\it -- Potentially Useful & \it 2,461 & \it 649 & \it 1,504+172+156 & \it 413  &  \it 456 &  \it 5,811 \\
\hline
\end{tabular}
\caption{Main statistics about the English CQA-QL corpus.}
\label{table:statistics:english}
\end{center}
\end{table*}

Each comment has a body text,\footnote{As most of the time the comment's subject is just ``RE: $<$question subject$>$'', we decided to drop it from the dataset.} as well as the following attributes:
\begin{itemize}
\itemsep0em 
\item RELC\_ID: comment identifier;
\item RELC\_USERID: identifier of the user posting the comment;
\item RELC\_USERNAME: name of the user posting the comment;
\item RELC\_RELEVANCE2ORGQ: human assessment about whether the comment is Good, Bad, or Potentially Useful with respect to the original question, OrgQuestion. This label can take one of the following values:
	\begin{itemize}
    \itemsep0em 
		\item \emph{Good}: at least one subquestion is directly answered by a portion of the comment;
     	\item \emph{PotentiallyUseful}: no subquestion is directly answered, but the comment gives potentially useful information about one or more subquestions (at test time, this class will be merged with Bad);
     	\item \emph{Bad}: no subquestion is answered and no useful information is provided (e.g., the answer is another question, a thanks, dialog with another user, a joke, irony, attack of other users, or is not in English, etc.).
	\end{itemize}
\item RELC\_RELEVANCE2RELQ: human assessment about whether the comment is Good, Bad, or PotentiallyUseful (again, the latter two are merged under Bad at test time) with respect to the related question, RelQuestion.
\end{itemize}

We used the CrowdFlower\footnote{http://www.crowdflower.com} crowdsourcing platform to annotate the gold labels for the three subtasks,
namely RELC\_RELEVANCE2RELQ for subtask A, RELQ\_RELEVANCE2ORGQ for subtask B, and RELC\_RELEVANCE2ORGQ for subtask C. 
We collected several annotations for each decision (there were at least three human annotators per example) and we resolved the discrepancies using the default mechanisms of CrowdFlower, which take into account the general quality of annotation for each annotator (based on the hidden tests). 

Unlike SemEval-2015 Task 3 \cite{nakov-EtAl:2015:SemEval}, where we excluded comments for which there was a lot of disagreement about the labels between the human annotators, this time we did not eliminate any comments (but we controlled the annotation quality with hidden tests), and thus we guarantee that for each question thread, we have the first ten comments without any comment being skipped.

To gather gold annotation labels, we created two annotation jobs on CrowdFlower, screenshots of which are shown in Figures~\ref{CFAnnotationFig} and \ref{CFAnnotation2Fig}.

The first annotation job aims to collect labels for subtasks B and C. We show a screenshot in Figure~\ref{CFAnnotationFig}. An annotation example consists of an original question, a related question, and the first ten comments for that related question. 

We asked the annotators to judge the relevance of the thread with respect to the original question (RELQ\_RELEVANCE2ORGQ, for subtask B), as well as the relevance of each comment with respect to the original question (RELC\_RELEVANCE2ORGQ, for subtask C).
Each example is judged by three annotators who must maintain 70\% accuracy throughout the job, measured on a hidden set of 121 examples.\footnote{The hidden tests for all subtasks were generated gradually. We started with a small number of initial tests, verified by two task coorganizers, and we gradually added more, choosing from those for which we had highest annotation agreement.} The average inter-annotator agreement on the training, development, and testing datasets is
80\%, 74\%, and 87\% for RELQ\_RELEVANCE2ORGQ,
and 83\%, 74\%, and 88\% for RELC\_RELEVANCE2ORGQ.

The second CrowdFlower job collects labels for subtask A; a screenshot is shown in Figure~\ref{CFAnnotation2Fig}. An annotation example consists of a question-comments thread, with ten comments, and we ask annotators to judge the relevance of each comment with respect to the thread question (RELC\_RELEVANCE2RELQ).
Again, each example is judged by three annotators who must maintain 70\% accuracy throughout the job, measured on a hidden set of 150 examples. The average inter-annotator agreement on the training, development, and testing datasets is
82\%, 89\%, and 79\% for RELC\_RELEVANCE2RELQ.

A fully annotated example is shown in Figure~\ref{EnglishXMLFig}.
Statistics about the datasets are shown in Table~\ref{table:statistics:english}.

Note that the training data is split into two parts, where part2 is noisier than part1. For part2, a different annotation setup was used,\footnote{Here are the annotation instructions we used for part2: http://alt.qcri.org/semeval2016/task3/data/uploads/annotation\_ instructions\_for\_part2.pdf} which confused the annotators, and they often provided annotation for RELC\_RELEVANCE2ORGQ while wrongly thinking that they were actually annotating RELC\_RELEVANCE2RELQ. Note that the development data was annotated with the same setup as training part2; however, we manually double-checked and corrected it. Instead, the training part1 and testing datasets used the less confusing, and thus higher-quality annotation setup described above.

Note also that in addition to the above-described canonical XML format, we further released the data in an alternative uncleansed\footnote{In fact, minimally cleansed, so that the XML file is valid.} multi-line format. We further released a simplified file format containing only the relevant information for subtask A, where duplicated related questions are removed.\footnote{The same question can be retrieved as related for different original questions. These are not repetitions for subtasks B and C, but they are such for subtask A.}
Finally, we reformatted the training, development, and test data from SemEval-2015 Task 3 \cite{nakov-EtAl:2015:SemEval}, to match the subtask A format for this year.

We released this reformatted SemEval-2015 Task 3, subtask A data as additional training data.
We further released a large unannotated dataset from Qatar Living with 189,941 questions and 1,894,456 comments, which is useful for unsupervised learning or for training domain-specific word embeddings.

\subsection{Arabic Dataset}
\label{sec:arabicdataset}

While at SemEval-2015 \cite{nakov-EtAl:2015:SemEval} we used a dataset from the Fatwa website, this year we 
changed the domain to medical, which is largely ignored for Arabic. We will refer to the Arabic corpus as CQA-MD.
We extracted data from three popular Arabic medical websites that allow visitors to post questions related to health and medical conditions, and to get answers by professional doctors.
We collected 1,531 question-answer (QA) pairs from WebTeb,\footnote{http://www.webteb.com/} 69,582 pairs from Al-Tibbi,\footnote{http://www.altibbi.com/} and 31,714 pairs from the medical corner of Islamweb.\footnote{http://consult.islamweb.net/}

We used the 1,531 questions from WebTeb as our original questions, and we looked to find related QA pairs from the other two websites.
We collected over 100,000 QA pairs in total from the other two websites, we indexed them in Solr, and we searched them trying to find answers to the WebTeb questions. 

\begin{figure*}[tbh]
\centering
\includegraphics[width=\textwidth, frame]{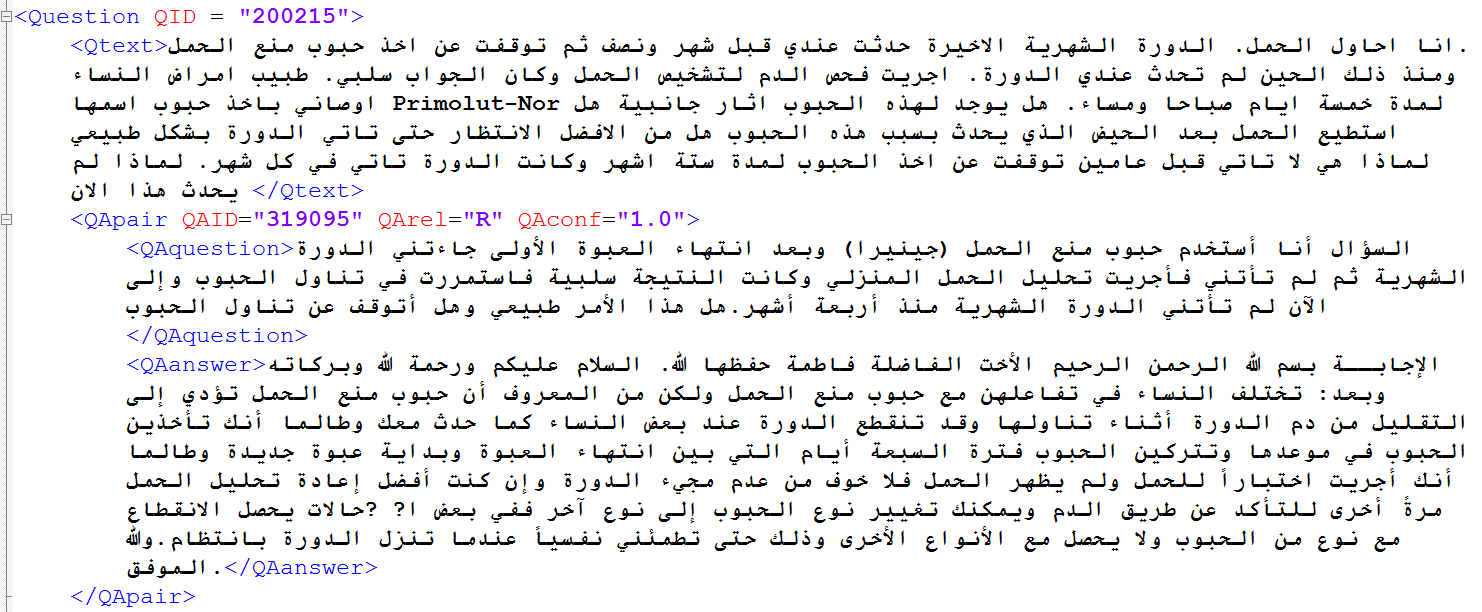}
\caption{Annotated question from the Arabic CQA-MD corpus.}
\label{fig:arabicsample}
\end{figure*}

We used several different query/document formulations to perform 21 retrieval runs, and we merged the retrieved results, ranking them according to the reciprocal rank fusion algorithm~\cite{cormack2009reciprocal}.
Finally, we truncated the result list to the 30 top-ranked QA pairs, ending up with 45,164 QA pairs\footnote{We had less than 30 answers for some questions.} for the 1,531 original questions.
Next, we used CrowdFlower to obtain judgments about the relevance of these QA pairs with respect to the original question using the following labels:
\begin{itemize}
\item ``\textbf{D}'' (Direct): The QA pair contains a direct answer to the original question such that if the user is searching for an answer to the original question, the proposed QA pair would be satisfactory and there would be no need to search any further.
\item ``\textbf{R}'' (Related): The QA pair contains an answer to the original question that covers some of the aspects raised in the original question, but this is not sufficient to answer it directly. With this QA pair, it would be expected that the user will continue the search to find a direct answer or more information.
\item ``\textbf{I}'' (Irrelevant): The QA pair contains an answer that is irrelevant to the original question.
\end{itemize}

We controlled the quality of annotation using a hidden set of 50 test questions. We had three judgments per example, which we combined using the CrowdFlower mechanism.
The average inter-annotator agreement was 81\%.

Finally, we divided the data into training, development and testing datasets, based on confidence, where the examples in the test dataset were those with the highest annotation confidence. We further double-checked and manually corrected some of the annotations for the development and the testing datasets whenever necessary. 

Figure~\ref{fig:arabicsample} shows part of the XML file we generated. We can see that, unlike the English data, there are no threads here, just a set of question-answer pairs; moreover, we do not provide much meta data, but we give information about the confidence of annotation (for the training and development datasets only, but not for the test dataset).

Table~\ref{table:statistics:arabic} shows some statistics about the dataset size and the distribution of the three classes in the CQA-MD corpus.

\begin{table}[tbh]
\small
\begin{center}
\begin{tabular}{lrrrr}
\bf Category & \bf Train & \bf Dev & \bf Test & \bf Total\\
\hline
\hline
\bf Questions & \bf 1,031 & \bf 250 & \bf 250 & \bf 1,531\\
\bf QA Pairs &  \bf 30,411 &  \bf 7,384 & \bf 7,369 & \bf 45,164\\
\it -- Direct & \it 917 & \it 70 & \it 65 & \it 1,052\\
\it -- Related & \it 17,412 & \it 1,446 & \it 1,353 & \it 20,211\\
\it -- Irrelevant & \it 12,082 & \it 5,868 & \it 5,951 & \it 23,901\\
\hline
\end{tabular}
\caption{Main statistics about the CQA-MD corpus.}
\label{table:statistics:arabic}
\end{center}
\end{table}


\section{Scoring}
\label{sec:scoring}

The official evaluation measure 
we used to rank the participating systems is Mean Average Precision (MAP) calculated for the ten comments a participating system has ranked highest. It is a well-established in Information Retrieval. We further report the results for two unofficial ranking measures, which we also calculate for the top-10 results only: Mean Reciprocal Rank (MRR) and Average Recall (AvgRec).
Additionally, we report the results for four standard classification measures, 
which we calculate over the full list of results: Precision, Recall, F$_1$ (with respect to the Good/Relevant class) and Accuracy.

We released a specialized scorer that calculates and reports all above-mentioned seven scores.

\section{Participants and Results}
\label{sec:results}

The list of all participating teams can be found in Table~\ref{table:teams}. The results for subtasks A, B, C, and D are shown in Tables \ref{table:results:subtaskA}, \ref{table:results:subtaskB}, \ref{table:results:subtaskC}, and \ref{table:results:subtaskD}, respectively. In all tables, the systems are ranked by the official MAP scores for their primary runs\footnote{Participants could submit one primary run, to be used for the official ranking, and up to two contrastive runs, which are scored but have unofficial status.} (shown in the third column). The following columns show the scores based on the other six unofficial measures; the ranking with respect to these additional measures are marked with a subindex (for the primary runs).

Eighteen teams participated in the challenge presenting a variety of approaches and features to address the different subtasks. They submitted a total of 95 runs (38 primary and 57 contrastive), which are broken down by subtasks in the following way: The English subtasks A, B and C attracted 12, 11, and 10 systems and 29, 25 and 28 runs, respectively. The Arabic subtask D got 5 systems and 13 runs.
The best MAP scores varied from 45.83 to 79.19, depending on the subtask. The best systems in each subtask were able to beat the baselines we provided by sizeable margins.



\begin{table*}[tbh]
\begin{center}
\begin{tabular}{clrrrrrrr}
& \bf Submission & \bf MAP & \bf \scriptsize AvgRec & \bf \scriptsize MRR & \bf \scriptsize P & \bf \scriptsize R & \bf \scriptsize F$_1$ & \bf \scriptsize Acc\\
\hline
\bf 1 & \bf Kelp-primary & \bf 79.19$_{1}$ & \bf \scriptsize 88.82$_{1}$ & \bf \scriptsize 86.42$_{1}$ & \bf \scriptsize 76.96$_{1}$ & \bf \scriptsize 55.30$_{8}$ & \bf \scriptsize 64.36$_{5}$ & \bf \scriptsize 75.11$_{2}$ \\
& ConvKN-contrastive1 & 78.71 & \scriptsize 88.98 & \scriptsize 86.15 & \scriptsize 77.78 & \scriptsize 53.72 & \scriptsize 63.55 & \scriptsize 74.95 \\
& SUper\_team-contrastive1 & 77.68 & \scriptsize 88.06 & \scriptsize 84.76 & \scriptsize 75.59 & \scriptsize 55.00 & \scriptsize 63.68 & \scriptsize 74.50 \\
\bf 2 & \bf ConvKN-primary & \bf 77.66$_{2}$ & \bf \scriptsize 88.05$_{3}$ & \bf \scriptsize 84.93$_{4}$ & \bf \scriptsize 75.56$_{2}$ & \bf \scriptsize 58.84$_{6}$ & \bf \scriptsize 66.16$_{2}$ & \bf \scriptsize 75.54$_{1}$ \\
\bf 3 & \bf SemanticZ-primary & \bf 77.58$_{3}$ & \bf \scriptsize 88.14$_{2}$ & \bf \scriptsize 85.21$_{2}$ & \bf \scriptsize 74.13$_{4}$ & \bf \scriptsize 53.05$_{10}$ & \bf \scriptsize 61.84$_{8}$ & \bf \scriptsize 73.39$_{5}$ \\
& ConvKN-contrastive2 & 77.29 & \scriptsize 87.77 & \scriptsize 85.03 & \scriptsize 74.74 & \scriptsize 59.67 & \scriptsize 66.36 & \scriptsize 75.41 \\
\bf 4 & \bf ECNU-primary & \bf 77.28$_{4}$ & \bf \scriptsize 87.52$_{5}$ & \bf \scriptsize 84.09$_{6}$ & \bf \scriptsize 70.46$_{6}$ & \bf \scriptsize 63.36$_{4}$ & \bf \scriptsize 66.72$_{1}$ & \bf \scriptsize 74.31$_{4}$ \\
& SemanticZ-contrastive1 & 77.16 & \scriptsize 87.73 & \scriptsize 84.08 & \scriptsize 75.29 & \scriptsize 53.20 & \scriptsize 62.35 & \scriptsize 73.88 \\
\bf 5 & \bf SUper\_team-primary & \bf 77.16$_{5}$ & \bf \scriptsize 87.98$_{4}$ & \bf \scriptsize 84.69$_{5}$ & \bf \scriptsize 74.43$_{3}$ & \bf \scriptsize 56.73$_{7}$ & \bf \scriptsize 64.39$_{4}$ & \bf \scriptsize 74.50$_{3}$ \\
& MTE-NN-contrastive2 & 76.98 & \scriptsize 86.98 & \scriptsize 85.50 & \scriptsize 58.71 & \scriptsize 70.28 & \scriptsize 63.97 & \scriptsize 67.83 \\
& SUper\_team-contrastive2 & 76.97 & \scriptsize 87.89 & \scriptsize 84.58 & \scriptsize 74.31 & \scriptsize 56.36 & \scriptsize 64.10 & \scriptsize 74.34 \\
& MTE-NN-contrastive1 & 76.86 & \scriptsize 87.03 & \scriptsize 84.36 & \scriptsize 55.84 & \scriptsize 77.35 & \scriptsize 64.86 & \scriptsize 65.93 \\
& SLS-contrastive2 & 76.71 & \scriptsize 87.17 & \scriptsize 84.38 & \scriptsize 59.45 & \scriptsize 67.95 & \scriptsize 63.41 & \scriptsize 68.13 \\
& SLS-contrastive1 & 76.46 & \scriptsize 87.47 & \scriptsize 83.27 & \scriptsize 60.09 & \scriptsize 69.68 & \scriptsize 64.53 & \scriptsize 68.87 \\
\bf 6 & \bf MTE-NN-primary & \bf 76.44$_{6}$ & \bf \scriptsize 86.74$_{7}$ & \bf \scriptsize 84.97$_{3}$ & \bf \scriptsize 56.28$_{9}$ & \bf \scriptsize 76.22$_{1}$ & \bf \scriptsize 64.75$_{3}$ & \bf \scriptsize 66.27$_{8}$ \\
\bf 7 & \bf SLS-primary & \bf 76.33$_{7}$ & \bf \scriptsize 87.30$_{6}$ & \bf \scriptsize 82.99$_{7}$ & \bf \scriptsize 60.36$_{8}$ & \bf \scriptsize 67.72$_{3}$ & \bf \scriptsize 63.83$_{6}$ & \bf \scriptsize 68.81$_{7}$ \\
& ECNU-contrastive2 & 75.71 & \scriptsize 86.14 & \scriptsize 82.53 & \scriptsize 63.60 & \scriptsize 66.67 & \scriptsize 65.10 & \scriptsize 70.95 \\
& SemanticZ-contrastive2 & 75.41 & \scriptsize 86.51 & \scriptsize 82.52 & \scriptsize 73.19 & \scriptsize 50.11 & \scriptsize 59.49 & \scriptsize 72.26 \\
& ICRC­-HIT-contrastive1 & 73.34 & \scriptsize 84.81 & \scriptsize 79.65 & \scriptsize 63.43 & \scriptsize 69.30 & \scriptsize 66.24 & \scriptsize 71.28 \\
\bf 8 & \bf ITNLP­-AiKF-primary & \bf 71.52$_{8}$ & \bf \scriptsize 82.67$_{9}$ & \bf \scriptsize 80.26$_{8}$ & \bf \scriptsize 73.18$_{5}$ & \bf \scriptsize 19.71$_{12}$ & \bf \scriptsize 31.06$_{12}$ & \bf \scriptsize 64.43$_{9}$ \\
& ECNU-contrastive1 & 71.34 & \scriptsize 83.39 & \scriptsize 78.62 & \scriptsize 66.95 & \scriptsize 41.31 & \scriptsize 51.09 & \scriptsize 67.86 \\
\bf 9 & \bf ICRC­-HIT-primary & \bf 70.90$_{9}$ & \bf \scriptsize 83.36$_{8}$ & \bf \scriptsize 77.38$_{10}$ & \bf \scriptsize 62.48$_{7}$ & \bf \scriptsize 62.53$_{5}$ & \bf \scriptsize 62.50$_{7}$ & \bf \scriptsize 69.51$_{6}$ \\
\bf 10 & \bf PMI-cool-primary & \bf 68.79$_{10}$ & \bf \scriptsize 79.94$_{10}$ & \bf \scriptsize 80.00$_{9}$ & \bf \scriptsize 47.81$_{12}$ & \bf \scriptsize 70.58$_{2}$ & \bf \scriptsize 57.00$_{9}$ & \bf \scriptsize 56.73$_{12}$ \\
& UH-PRHLT-contrastive1 & 67.57 & \scriptsize 79.50 & \scriptsize 77.08 & \scriptsize 54.10 & \scriptsize 50.11 & \scriptsize 52.03 & \scriptsize 62.45 \\
\bf 11 & \bf UH-PRHLT-primary & \bf 67.42$_{11}$ & \bf \scriptsize 79.38$_{11}$ & \bf \scriptsize 76.97$_{11}$ & \bf \scriptsize 55.64$_{10}$ & \bf \scriptsize 46.80$_{11}$ & \bf \scriptsize 50.84$_{11}$ & \bf \scriptsize 63.21$_{10}$ \\
& UH-PRHLT-contrastive2 & 67.33 & \scriptsize 79.34 & \scriptsize 76.73 & \scriptsize 54.97 & \scriptsize 49.13 & \scriptsize 51.89 & \scriptsize 62.97 \\
\bf 12 & \bf QAIIIT-primary & \bf 62.24$_{12}$ & \bf \scriptsize 75.41$_{12}$ & \bf \scriptsize 70.58$_{12}$ & \bf \scriptsize 50.28$_{11}$ & \bf \scriptsize 53.50$_{9}$ & \bf \scriptsize 51.84$_{10}$ & \bf \scriptsize 59.60$_{11}$ \\
& QAIIIT-contrastive2 & 61.93 & \scriptsize 75.22 & \scriptsize 69.95 & \scriptsize 49.48 & \scriptsize 49.96 & \scriptsize 49.72 & \scriptsize 58.93 \\
& QAIIIT-contrastive1 & 61.80 & \scriptsize 75.12 & \scriptsize 69.76 & \scriptsize 49.85 & \scriptsize 50.94 & \scriptsize 50.39 & \scriptsize 59.24 \\
\hline
 & Baseline 1 (chronological) & \bf 59.53 & \scriptsize \bf 72.60 & \scriptsize \bf 67.83 & \scriptsize --- & \scriptsize --- & \scriptsize --- & \scriptsize --- \\
 & Baseline 2 (random) & 52.80 & \scriptsize 66.52 & \scriptsize 58.71 &  \scriptsize 40.56 & \scriptsize 74.57 & \scriptsize 52.55 & \scriptsize 45.26 \\
 & Baseline 3 (all `true') & --- & \scriptsize --- & \scriptsize --- &  \scriptsize 40.64 & \scriptsize 100.00 & \scriptsize \bf 57.80 & \scriptsize 40.64 \\
 & Baseline 4 (all `false') & --- & \scriptsize --- & \scriptsize --- &  \scriptsize --- & \scriptsize --- & \scriptsize \bf --- & \scriptsize \bf 59.36 \\
\hline
\end{tabular}
\caption{\textbf{Subtask A, English (Question-Comment Similarity):} results for all submissions. The first column shows the rank of the primary runs with respect to the official MAP score. The second column contains the team's name and its submission type (primary vs. contrastive).
The following columns show the results for the primary, and then for other, unofficial evaluation measures. The subindices show the rank of the primary runs with respect to the evaluation measure in the respective column.
}
\label{table:results:subtaskA}
\end{center}
\end{table*}

\subsection{Subtask A, English (Question-Comment Similarity)}

Table~\ref{table:results:subtaskA} shows the results for subtask A, English, which attracted 12 teams, which submitted 29 runs: 12 primary and 17 contrastive. 
The last four rows of the table show the performance of four baselines.
The first one is the chronological ranking, where the comments are ordered by their time of posting; we can see that all submissions outperform this baseline on all three ranking measures.
The second baseline is a random baseline, which outperforms some systems in terms of F$_1$, primarily because of having very high Recall.
Baseline 3 classifies all comments as Good, and it outperforms four of the primary systems in terms of F$_1$.
Finally, baseline 4 classifies all comments as Bad; it outperforms one of the primary systems in terms of Accuracy.

The winning team is that of KeLP \cite{SemEval2016:task3:KeLP}, which achieved the highest MAP of 79.19, outperforming the second best by a margin; they are also first on AvgRec and MRR, and second on Accuracy. They learn semantic relations between questions and answers using kernels and previously-proposed features from \cite{barroncedeno-EtAl:2015:ACL-IJCNLP}. Their system is based on the KeLP machine learning platform \cite{filice-EtAl:2015:KeLP2015}, and thus the name of the team.

The second best system is that of ConvKN \cite{SemEval2016:task3:ConvKN} with MAP of 77.66; it is also first on Accuracy, second on F$_1$, and third on AvgRec. The system combines convolutional tree kernels and convolutional neural networks, together with text similarity and thread-specific features. Their contrastive1 run achieved even better results: MAP of 78.71.

The third best system is SemanticZ \cite{SemEval2016:task3:SemanticZ} with MAP of 77.58. They use semantic similarity based on word embeddings and topics; they are second on AvgRec and MRR.

Note also the cluster of systems of very close MAP: ConvKN \cite{SemEval2016:task3:ConvKN} with 77.66, SemanticZ \cite{SemEval2016:task3:SemanticZ} with 77.58, ECNU \cite{SemEval2016:task3:ECNU} with 77.28, and SUper\_team \cite{SemEval2016:task3:SUper} with 77.16. The latter also has a contrastive run with MAP of 77.68, which would have ranked second.

\begin{table*}[tbh]
\begin{center}
\begin{tabular}{clrrrrrrr}
& \bf Submission & \bf MAP & \bf \scriptsize AvgRec & \bf \scriptsize MRR & \bf \scriptsize P & \bf \scriptsize R & \bf \scriptsize F$_1$ & \bf \scriptsize Acc\\
\hline
& UH-PRHLT-contrastive2 & 77.33 & \scriptsize 90.84 & \scriptsize 83.93 & \scriptsize 63.57 & \scriptsize 70.39 & \scriptsize 66.80 & \scriptsize 76.71 \\
\bf 1 & \bf UH-PRHLT-primary & \bf 76.70$_{1}$ & \bf \scriptsize 90.31$_{4}$ & \bf \scriptsize 83.02$_{4}$ & \bf \scriptsize 63.53$_{7}$ & \bf \scriptsize 69.53$_{3}$ & \bf \scriptsize 66.39$_{3}$ & \bf \scriptsize 76.57$_{4}$ \\
& UH-PRHLT-contrastive1 & 76.56 & \scriptsize 90.22 & \scriptsize 83.02 & \scriptsize 62.74 & \scriptsize 70.82 & \scriptsize 66.53 & \scriptsize 76.29 \\
& Kelp-contrastive1 & 76.28 & \scriptsize 91.33 & \scriptsize 82.71 & \scriptsize 63.83 & \scriptsize 77.25 & \scriptsize 69.90 & \scriptsize 77.86 \\
& Kelp-contrastive2 & 76.27 & \scriptsize 91.44 & \scriptsize 84.10 & \scriptsize 64.06 & \scriptsize 77.25 & \scriptsize 70.04 & \scriptsize 78.00 \\
& SLS-contrastive1 & 76.17 & \scriptsize 90.55 & \scriptsize 85.48 & \scriptsize 74.39 & \scriptsize 52.36 & \scriptsize 61.46 & \scriptsize 78.14 \\
& SLS-contrastive2 & 76.09 & \scriptsize 90.14 & \scriptsize 84.21 & \scriptsize 77.21 & \scriptsize 45.06 & \scriptsize 56.91 & \scriptsize 77.29 \\
\bf 2 & \bf ConvKN-primary & \bf 76.02$_{2}$ & \bf \scriptsize 90.70$_{2}$ & \bf \scriptsize 84.64$_{1}$ & \bf \scriptsize 68.58$_{3}$ & \bf \scriptsize 66.52$_{6}$ & \bf \scriptsize 67.54$_{2}$ & \bf \scriptsize 78.71$_{3}$ \\
\bf 3 & \bf Kelp-primary & \bf 75.83$_{3}$ & \bf \scriptsize 91.02$_{1}$ & \bf \scriptsize 82.71$_{6}$ & \bf \scriptsize 66.79$_{4}$ & \bf \scriptsize 75.97$_{2}$ & \bf \scriptsize 71.08$_{1}$ & \bf \scriptsize 79.43$_{1}$ \\
& ConvKN-contrastive1 & 75.57 & \scriptsize 89.64 & \scriptsize 83.57 & \scriptsize 63.77 & \scriptsize 72.53 & \scriptsize 67.87 & \scriptsize 77.14 \\
\bf 4 & \bf SLS-primary & \bf 75.55$_{4}$ & \bf \scriptsize 90.65$_{3}$ & \bf \scriptsize 84.64$_{1}$ & \bf \scriptsize 76.33$_{2}$ & \bf \scriptsize 55.36$_{9}$ & \bf \scriptsize 64.18$_{6}$ & \bf \scriptsize 79.43$_{1}$ \\
& SUper\_team-contrastive1 & 75.17 & \scriptsize 88.84 & \scriptsize 83.66 & \scriptsize 63.25 & \scriptsize 63.52 & \scriptsize 63.38 & \scriptsize 75.57 \\
\bf 5 & \bf ICL00-primary & \bf 75.11$_{5}$ & \bf \scriptsize 89.33$_{5}$ & \bf \scriptsize 83.02$_{4}$ & \bf \scriptsize 33.29$_{11}$ & \bf \scriptsize 100.00$_{1}$ & \bf \scriptsize 49.95$_{9}$ & \bf \scriptsize 33.29$_{11}$ \\
& ICL00-contrastive1 & 74.89 & \scriptsize 89.08 & \scriptsize 82.71 & \scriptsize 33.29 & \scriptsize 100.00 & \scriptsize 49.95 & \scriptsize 33.29 \\
\bf 6 & \bf SUper\_team-primary & \bf 74.82$_{6}$ & \bf \scriptsize 88.54$_{7}$ & \bf \scriptsize 83.66$_{3}$ & \bf \scriptsize 63.64$_{6}$ & \bf \scriptsize 57.08$_{8}$ & \bf \scriptsize 60.18$_{7}$ & \bf \scriptsize 74.86$_{7}$ \\
& ICL00-contrastive2 & 74.05 & \scriptsize 89.11 & \scriptsize 82.79 & \scriptsize 33.29 & \scriptsize 100.00 & \scriptsize 49.95 & \scriptsize 33.29 \\
\bf 7 & \bf ECNU-primary & \bf 73.92$_{7}$ & \bf \scriptsize 89.07$_{6}$ & \bf \scriptsize 81.48$_{7}$ & \bf \scriptsize 100.00$_{1}$ & \bf \scriptsize 18.03$_{11}$ & \bf \scriptsize 30.55$_{11}$ & \bf \scriptsize 72.71$_{9}$ \\
& ECNU-contrastive1 & 73.25 & \scriptsize 88.55 & \scriptsize 80.81 & \scriptsize 100.00 & \scriptsize 18.03 & \scriptsize 30.55 & \scriptsize 72.71 \\
& ECNU-contrastive2 & 71.62 & \scriptsize 86.55 & \scriptsize 80.88 & \scriptsize 54.61 & \scriptsize 71.24 & \scriptsize 61.82 & \scriptsize 70.71 \\
\bf 8 & \bf ITNLP­-AiKF-primary & \bf 71.43$_{8}$ & \bf \scriptsize 87.31$_{8}$ & \bf \scriptsize 81.28$_{8}$ & \bf \scriptsize 62.75$_{9}$ & \bf \scriptsize 68.67$_{4}$ & \bf \scriptsize 65.57$_{4}$ & \bf \scriptsize 76.00$_{6}$ \\
\bf 9 & \bf UniMelb-primary & \bf 70.20$_{9}$ & \bf \scriptsize 86.21$_{9}$ & \bf \scriptsize 78.58$_{11}$ & \bf \scriptsize 63.96$_{5}$ & \bf \scriptsize 54.08$_{10}$ & \bf \scriptsize 58.60$_{8}$ & \bf \scriptsize 74.57$_{8}$ \\
\bf 10 & \bf overfitting-primary & \bf 69.68$_{10}$ & \bf \scriptsize 85.10$_{10}$ & \bf \scriptsize 80.18$_{9}$ & \bf \scriptsize 63.20$_{8}$ & \bf \scriptsize 67.81$_{5}$ & \bf \scriptsize 65.42$_{5}$ & \bf \scriptsize 76.14$_{5}$ \\
& QAIIIT-contrastive1 & 69.24 & \scriptsize 85.24 & \scriptsize 80.30 & \scriptsize 38.99 & \scriptsize 66.09 & \scriptsize 49.04 & \scriptsize 54.29 \\
\bf 11 & \bf QAIIIT-primary & \bf 69.04$_{11}$ & \bf \scriptsize 84.53$_{11}$ & \bf \scriptsize 79.55$_{10}$ & \bf \scriptsize 39.53$_{10}$ & \bf \scriptsize 64.81$_{7}$ & \bf \scriptsize 49.11$_{10}$ & \bf \scriptsize 55.29$_{10}$ \\
& QAIIIT-contrastive2 & 46.23 & \scriptsize 68.07 & \scriptsize 48.92 & \scriptsize 36.25 & \scriptsize 51.50 & \scriptsize 42.55 & \scriptsize 53.71 \\
\hline
 & Baseline 1 (IR) & \bf 74.75 & \scriptsize \bf 88.30 & \scriptsize \bf 83.79 & \scriptsize --- & \scriptsize --- & \scriptsize --- & \scriptsize --- \\
 & Baseline 2 (random) & 46.98 & \scriptsize 67.92 & \scriptsize 50.96  & \scriptsize 32.58 & \scriptsize 73.82 & \scriptsize 45.20 & \scriptsize 40.43 \\
 & Baseline 3 (all `true') & --- & \scriptsize --- & \scriptsize --- & \scriptsize 33.29 & \scriptsize 100.00 & \scriptsize \bf 49.95 & \scriptsize 33.29 \\
 & Baseline 4 (all `false') & --- & \scriptsize --- & \scriptsize --- & \scriptsize --- & \scriptsize --- & \scriptsize --- & \scriptsize \bf 66.71 \\
 \hline
\end{tabular}
\caption{\textbf{Subtask B, English (Question-Question Similarity):} results for all submissions. The first column shows the rank of the primary runs with respect to the official MAP score. The second column contains the team's name and its submission type (primary vs. contrastive).
The following columns show the results for the primary, and then for other, unofficial evaluation measures. The subindices show the rank of the primary runs with respect to the evaluation measure in the respective column.
}
\label{table:results:subtaskB}
\end{center}
\end{table*}

\subsection{Subtask B, English (Question-Question Similarity)}

Table~\ref{table:results:subtaskB} shows the results for subtask B, English, which attracted 11 teams and 25 runs: 11 primary and 14 contrastive. This turns out to be a hard task. For example, the IR baseline (i.e., ordering the related questions in the order provided by the search engine) outperforms 5 of the 11 systems in terms of MAP; it also outperforms several systems in terms of MRR and AvgRec.
The random baseline outperforms one system in terms of F$_1$ and Accuracy, again due to high recall.
The all-Good baseline outperforms two systems on F$_1$,
while the all-Bad baseline outperforms two systems on Accuracy.

The winning team is that of UH-PRHLT \cite{SemEval2016:task3:UH-PRHLT}, which achieved MAP of 76.70 (just 2 MAP points over the IR baseline). They use distributed representations of words, knowledge graphs generated with BabelNet, and frames from FrameNet.
Their contrastive2 run is even better, with MAP of 77.33.

The second best system is that of ConvKN \cite{SemEval2016:task3:ConvKN} with MAP of 76.02; they are also first on MRR, second on AvgRec and F$_1$, and third on Accuracy.

The third best system is KeLP \cite{SemEval2016:task3:KeLP} with MAP of 75.83; they are also first on AvgRec, F$_1$, and Accuracy. They have a contrastive run with MAP of 76.28, which would have ranked second.

The fourth best, SLS \cite{SemEval2016:task3:SLS} is very close, with MAP of 75.55; it is also first on MRR and Accuracy, and third on AvgRec. It uses a bag-of-vectors approach with various vector- and text-based features, and different neural network approaches including CNNs and LSTMs to capture the semantic similarity between questions and answers.

\begin{table*}[tbh]
\begin{center}
\begin{tabular}{clrrrrrrr}
& \bf Submission & \bf MAP & \bf \scriptsize AvgRec & \bf \scriptsize MRR & \bf \scriptsize P & \bf \scriptsize R & \bf \scriptsize F$_1$ & \bf \scriptsize Acc\\
\hline
& Kelp-contrastive2 & 55.58 & \scriptsize 63.36 & \scriptsize 61.19 & \scriptsize 32.21 & \scriptsize 70.18 & \scriptsize 44.16 & \scriptsize 83.41 \\
\bf 1 & \bf SUper\_team-primary & \bf 55.41$_{1}$ & \bf \scriptsize 60.66$_{1}$ & \bf \scriptsize 61.48$_{1}$ & \bf \scriptsize 18.03$_{7}$ & \bf \scriptsize 63.15$_{4}$ & \bf \scriptsize 28.05$_{4}$ & \bf \scriptsize 69.73$_{8}$ \\
& SUper\_team-contrastive2 & 53.48 & \scriptsize 59.40 & \scriptsize 59.09 & \scriptsize 18.42 & \scriptsize 66.97 & \scriptsize 28.89 & \scriptsize 69.20 \\
\bf 2 & \bf Kelp-primary & \bf 52.95$_{2}$ & \bf \scriptsize 59.27$_{2}$ & \bf \scriptsize 59.23$_{2}$ & \bf \scriptsize 33.63$_{5}$ & \bf \scriptsize 64.53$_{3}$ & \bf \scriptsize 44.21$_{1}$ & \bf \scriptsize 84.79$_{5}$ \\
& Kelp-contrastive1 & 52.95 & \scriptsize 59.34 & \scriptsize 58.06 & \scriptsize 34.08 & \scriptsize 65.29 & \scriptsize 44.78 & \scriptsize 84.96 \\
\bf 3 & \bf SemanticZ-primary & \bf 51.68$_{3}$ & \bf \scriptsize 53.43$_{6}$ & \bf \scriptsize 55.96$_{4}$ & \bf \scriptsize 17.11$_{8}$ & \bf \scriptsize 57.65$_{5}$ & \bf \scriptsize 26.38$_{5}$ & \bf \scriptsize 69.94$_{7}$ \\
& SemanticZ-contrastive1 & 51.46 & \scriptsize 52.69 & \scriptsize 55.75 & \scriptsize 16.94 & \scriptsize 57.49 & \scriptsize 26.17 & \scriptsize 69.69 \\
& MTE-NN-contrastive2 & 49.49 & \scriptsize 55.78 & \scriptsize 51.80 & \scriptsize 15.68 & \scriptsize 73.09 & \scriptsize 25.82 & \scriptsize 60.76 \\
\bf 4 & \bf MTE-NN-primary & \bf 49.38$_{4}$ & \bf \scriptsize 55.44$_{4}$ & \bf \scriptsize 51.56$_{7}$ & \bf \scriptsize 15.26$_{9}$ & \bf \scriptsize 76.15$_{2}$ & \bf \scriptsize 25.43$_{6}$ & \bf \scriptsize 58.27$_{9}$ \\
\bf 5 & \bf ICL00-primary & \bf 49.19$_{5}$ & \bf \scriptsize 51.07$_{7}$ & \bf \scriptsize 53.89$_{6}$ & \bf \scriptsize  9.34$_{10}$ & \bf \scriptsize 100.00$_{1}$ & \bf \scriptsize 17.09$_{8}$ & \bf \scriptsize  9.34$_{10}$ \\
\bf 6 & \bf SLS-primary & \bf 49.09$_{6}$ & \bf \scriptsize 56.04$_{3}$ & \bf \scriptsize 55.98$_{3}$ & \bf \scriptsize 47.85$_{2}$ & \bf \scriptsize 13.61$_{8}$ & \bf \scriptsize 21.19$_{7}$ & \bf \scriptsize 90.54$_{2}$ \\
& SemanticZ-contrastive2 & 48.76 & \scriptsize 50.72 & \scriptsize 53.85 & \scriptsize 16.92 & \scriptsize 57.34 & \scriptsize 26.13 & \scriptsize 69.71 \\
& MTE-NN-contrastive1 & 48.52 & \scriptsize 54.71 & \scriptsize 50.51 & \scriptsize 15.13 & \scriptsize 76.61 & \scriptsize 25.27 & \scriptsize 57.67 \\
\bf 7 & \bf ITNLP­-AiKF-primary & \bf 48.49$_{7}$ & \bf \scriptsize 55.16$_{5}$ & \bf \scriptsize 55.21$_{5}$ & \bf \scriptsize 30.05$_{6}$ & \bf \scriptsize 50.92$_{6}$ & \bf \scriptsize 37.80$_{2}$ & \bf \scriptsize 84.34$_{6}$ \\
& ECNU-contrastive1 & 48.49 & \scriptsize 53.17 & \scriptsize 53.47 & \scriptsize 68.75 & \scriptsize  8.41 & \scriptsize 14.99 & \scriptsize 91.09 \\
& SUper\_team-contrastive1 & 48.23 & \scriptsize 54.93 & \scriptsize 54.85 & \scriptsize 22.81 & \scriptsize 36.70 & \scriptsize 28.14 & \scriptsize 82.49 \\
& ECNU-contrastive2 & 47.24 & \scriptsize 53.21 & \scriptsize 51.89 & \scriptsize 70.27 & \scriptsize  7.95 & \scriptsize 14.29 & \scriptsize 91.09 \\
& ICL00-contrastive1 & 47.23 & \scriptsize 49.71 & \scriptsize 50.28 & \scriptsize  9.34 & \scriptsize 100.00 & \scriptsize 17.09 & \scriptsize  9.34 \\
\bf 8 & \bf ConvKN-primary & \bf 47.15$_{8}$ & \bf \scriptsize 47.46$_{10}$ & \bf \scriptsize 51.43$_{8}$ & \bf \scriptsize 45.97$_{3}$ & \bf \scriptsize  8.72$_{10}$ & \bf \scriptsize 14.65$_{10}$ & \bf \scriptsize 90.51$_{3}$ \\
& SLS-contrastive1 & 46.48 & \scriptsize 53.31 & \scriptsize 52.53 & \scriptsize 16.24 & \scriptsize 85.93 & \scriptsize 27.32 & \scriptsize 57.29 \\
\bf 9 & \bf ECNU-primary & \bf 46.47$_{9}$ & \bf \scriptsize 50.92$_{8}$ & \bf \scriptsize 51.41$_{9}$ & \bf \scriptsize 66.29$_{1}$ & \bf \scriptsize  9.02$_{9}$ & \bf \scriptsize 15.88$_{9}$ & \bf \scriptsize 91.07$_{1}$ \\
& SLS-contrastive2 & 46.39 & \scriptsize 52.83 & \scriptsize 51.17 & \scriptsize 16.18 & \scriptsize 85.63 & \scriptsize 27.22 & \scriptsize 57.23 \\
& UH-PRHLT-contrastive1 & 43.37 & \scriptsize 48.01 & \scriptsize 48.43 & \scriptsize 38.56 & \scriptsize 32.72 & \scriptsize 35.40 & \scriptsize 88.84 \\
& UH-PRHLT-contrastive2 & 43.32 & \scriptsize 47.97 & \scriptsize 48.45 & \scriptsize 38.21 & \scriptsize 32.72 & \scriptsize 35.26 & \scriptsize 88.77 \\
& ConvKN-contrastive1 & 43.31 & \scriptsize 44.19 & \scriptsize 48.89 & \scriptsize 30.00 & \scriptsize  3.21 & \scriptsize  5.80 & \scriptsize 90.26 \\
\bf 10 & \bf UH-PRHLT-primary & \bf 43.20$_{10}$ & \bf \scriptsize 47.96$_{9}$ & \bf \scriptsize 47.79$_{10}$ & \bf \scriptsize 37.65$_{4}$ & \bf \scriptsize 34.25$_{7}$ & \bf \scriptsize 35.87$_{3}$ & \bf \scriptsize 88.56$_{4}$ \\
& ICL00-contrastive2 & 41.32 & \scriptsize 44.56 & \scriptsize 43.55 & \scriptsize  9.34 & \scriptsize 100.00 & \scriptsize 17.09 & \scriptsize  9.34 \\
& ConvKN-contrastive2 & 41.12 & \scriptsize 38.89 & \scriptsize 44.17 & \scriptsize 33.55 & \scriptsize 32.11 & \scriptsize 32.81 & \scriptsize 87.71 \\
\hline
 & Baseline 1 (IR+chronological) & \bf 40.36 & \scriptsize \bf 45.97 & \scriptsize \bf 45.83 & \scriptsize --- & \scriptsize --- & \scriptsize --- & \scriptsize --- \\
 & Baseline 2 (random) & 15.01 & \scriptsize 11.44 & \scriptsize 15.19 & \scriptsize 9.40 & \scriptsize 75.69 & \scriptsize 16.73 & \scriptsize 29.59 \\
 & Baseline 3 (all `true') & --- & \scriptsize --- & \scriptsize --- & \scriptsize 9.34 & \scriptsize 100.00 & \scriptsize \bf 17.09 & \scriptsize 9.34 \\
 & Baseline 4 (all `false') & --- & \scriptsize --- & \scriptsize --- & \scriptsize --- & \scriptsize --- & \scriptsize --- & \scriptsize \bf 90.66 \\
 \hline
\end{tabular}
\caption{\textbf{Subtask C, English (Question-External Comment Similarity):} results for all submissions. The first column shows the rank of the primary runs with respect to the official MAP score. The second column contains the team's name and its submission type (primary vs. contrastive).
The following columns show the results for the primary, and then for other, unofficial evaluation measures. The subindices show the rank of the primary runs with respect to the evaluation measure in the respective column.
}
\label{table:results:subtaskC}\end{center}
\end{table*}

\subsection{Subtask C, English (Question-External Comment Similarity)}

The results for subtask C, English are shown in Table~\ref{table:results:subtaskC}. This subtask attracted 10 teams, and 28 runs: 10 primary and 18 contrastive. Here the teams performed much better than they did for subtask B.
The first three baselines were all outperformed by all participating systems.
However, due to severe class imbalance, the all-Bad baseline outperformed 9 out of the 10 participating teams in terms of Accuracy.

The best system in this subtask is that of the SUper\_team \cite{SemEval2016:task3:SUper}, which achieved MAP of 55.41; the system is also first on AvgRec and MRR. It used a rich set of features, grouped into three categories: question-specific features, answer-specific features, and question-answer similarity features. This includes more or less standard metadata, lexical, semantic, and user-related features, as well as some exotic ones such as features related to readability, credibility, as well as goodness polarity lexicons.\footnote{These goodness polarity lexicons were at the core of another system, PMI-cool \cite{SemEval2016:task3:PMI-cool}, which did not perform very well as it limited itself to lexicons and ignored other important features.} 
It is important to note that this system did not try to solve subtask C directly, but rather just multipled their predicted score for subtask A by the reciprocal rank of the related question in the list of related questions (as returned by the search engine, and as readily provided by the organizers as an attribute in the XML file) for the original question. In fact, this is not an isolated case, but an approach taken by several participants in subtask C.

The second best system is that of KeLP \cite{SemEval2016:task3:KeLP}, with MAP of 52.95; they are also first on F$_1$, and second on AvgRec and MRR. KeLP also has a contrastive run with a MAP of 55.58, which would have made them first.
This team really tried to solve the actual subtask C by means of stacking classifiers: they used their subtask A classifier to judge how good the answer is with respect to the original and with respect to the related question. Moreover, they used their subtask B classifier to judge the relatedness of the related question with respect to the original question. Finally, they used these three scores, together with some features based on them, to train a classifier that solves subtask C. 

In fact, we anticipated solutions like this when we designed the task, i.e., that participants would solve subtasks A and B, and use them as auxiliary tasks to attack the main task, namely subtask C. Unfortunately, subtask B turned out to be too hard, and thus many participants decided to skip it and just to use the search engine's reciprocal rank. 

The third best system is SemanticZ \cite{SemEval2016:task3:SemanticZ}, with MAP of 51.68. Similarly to SUper\_team, they simply multiply their predicted score for subtask A by the reciprocal rank of the related question in the list of related questions for the original question.

\begin{table*}[tbh]
\begin{center}
\begin{tabular}{clrrrrrrr}
& \bf Submission & \bf MAP & \bf \scriptsize AvgRec & \bf \scriptsize MRR & \bf \scriptsize P & \bf \scriptsize R & \bf \scriptsize F$_1$ & \bf \scriptsize Acc\\
\hline
 \bf 1 & \bf SLS-primary & \bf 45.83$_{1}$ & \bf \scriptsize 51.01$_{1}$ & \bf \scriptsize 53.66$_{1}$ & \bf \scriptsize 34.45$_{1}$ & \bf \scriptsize 52.33$_{3}$ & \bf \scriptsize 41.55$_{1}$ & \bf \scriptsize 71.67$_{1}$ \\
\bf 2 & \bf ConvKN-primary & \bf 45.50$_{2}$ & \bf \scriptsize 50.13$_{2}$ & \bf \scriptsize 52.55$_{2}$ & \bf \scriptsize 28.55$_{2}$ & \bf \scriptsize 64.53$_{2}$ & \bf \scriptsize 39.58$_{2}$ & \bf \scriptsize 62.10$_{3}$ \\
& SLS-contrastive1 & 44.94 & \scriptsize 49.72 & \scriptsize 51.58 & \scriptsize 62.96 & \scriptsize  2.40 & \scriptsize  4.62 & \scriptsize 80.95 \\
\bf 3 & \bf RDI\_team-primary & \bf 43.80$_{3}$ & \bf \scriptsize 47.45$_{3}$ & \bf \scriptsize 49.21$_{3}$ & \bf \scriptsize 19.24$_{5}$ & \bf \scriptsize 100.00$_{1}$ & \bf \scriptsize 32.27$_{4}$ & \bf \scriptsize 19.24$_{5}$ \\
& SLS-contrastive2 & 42.95 & \scriptsize 47.61 & \scriptsize 49.55 & \scriptsize 27.20 & \scriptsize 74.40 & \scriptsize 39.84 & \scriptsize 56.76 \\
& RDI\_team-contrastive1 & 42.18 & \scriptsize 47.03 & \scriptsize 47.93 & \scriptsize 19.24 & \scriptsize 100.00 & \scriptsize 32.27 & \scriptsize 19.24 \\
& ConvKN-contrastive2 & 39.98 & \scriptsize 43.68 & \scriptsize 46.41 & \scriptsize 26.26 & \scriptsize 68.39 & \scriptsize 37.95 & \scriptsize 57.00 \\
& QU-IR-contrastive2 & 39.07 & \scriptsize 42.72 & \scriptsize 44.14 & \scriptsize 24.90 & \scriptsize 44.08 & \scriptsize 31.82 & \scriptsize 63.66 \\
& RDI\_team-contrastive2 & 38.84 & \scriptsize 42.98 & \scriptsize 42.97 & \scriptsize 19.24 & \scriptsize 100.00 & \scriptsize 32.27 & \scriptsize 19.24 \\
\bf 4 & \bf QU-IR-primary & \bf 38.63$_{4}$ & \bf \scriptsize 44.10$_{4}$ & \bf \scriptsize 46.27$_{4}$ & \bf \scriptsize 25.50$_{3}$ & \bf \scriptsize 45.13$_{5}$ & \bf \scriptsize 32.59$_{3}$ & \bf \scriptsize 64.07$_{2}$ \\
& ConvKN-contrastive1 & 38.33 & \scriptsize 42.09 & \scriptsize 43.75 & \scriptsize 20.38 & \scriptsize 96.95 & \scriptsize 33.68 & \scriptsize 26.58 \\
& QU-IR-contrastive1 & 37.80 & \scriptsize 40.96 & \scriptsize 44.39 & \scriptsize 23.54 & \scriptsize 41.89 & \scriptsize 30.14 & \scriptsize 62.64 \\
\bf 5 & \bf UPC\_USMBA-primary & \bf 29.09$_{5}$ & \bf \scriptsize 30.04$_{5}$ & \bf \scriptsize 34.04$_{5}$ & \bf \scriptsize 20.14$_{4}$ & \bf \scriptsize 51.69$_{4}$ & \bf \scriptsize 28.99$_{5}$ & \bf \scriptsize 51.27$_{4}$ \\
\hline
 & Baseline 1 (chronological) & \bf 28.88 & \scriptsize \bf 28.71 & \scriptsize \bf 30.93 & \scriptsize --- & \scriptsize --- & \scriptsize --- & \scriptsize --- \\
 & Baseline 2 (random) & 29.79 & \scriptsize 31.00 & \scriptsize 33.71 & \scriptsize 19.53 & \scriptsize 20.66 & \scriptsize 20.08 & \scriptsize 68.35 \\
 & Baseline 3 (all `true') & --- & \scriptsize --- & \scriptsize --- & \scriptsize 19.24 & \scriptsize 100.00 & \scriptsize \bf 32.27 & \scriptsize 19.24 \\
 & Baseline 4 (all `false') & --- & \scriptsize --- & \scriptsize --- & \scriptsize --- & \scriptsize --- & \scriptsize --- & \scriptsize \bf 80.76 \\
 \hline
\end{tabular}
\caption{\textbf{Subtask D, Arabic (Reranking the correct answers for a new question):} results for all submissions. The first column shows the rank of the primary runs with respect to the official MAP score. The second column contains the team's name and its submission type (primary vs. contrastive).
The following columns show the results for the primary, and then for other, unofficial evaluation measures. The subindices show the rank of the primary runs with respect to the evaluation measure in the respective column.
}
\label{table:results:subtaskD}
\end{center}
\end{table*}

\subsection{Subtask D, Arabic (Reranking the correct answers for a new question)}

Finally, the results for subtask D, Arabic are shown in Table~\ref{table:results:subtaskD}. It attracted 5 teams, which submitted 13 runs: 5 primary and 8 contrastive. 
As the class imbalance here is even more severe than for subtask C, the all-Bad baseline outperforms all participating systems in terms of Accuracy.
In contrast, the all-Good baseline only outperforms one system in terms of F$_1$.
Here the teams perform much better than for subtask B.
The random baseline outperforms one system in terms of both MAP and AvgRec.

The clear winner here is SLS \cite{SemEval2016:task3:SLS}, which is ranked first on all measures: MAP, AvgRec, MRR, F$_1$, and Accuracy. Yet, their MAP of 45.83 is only slightly better than that of ConvKN, 45.50, which ranks second on MAP, AvgRec, MRR and F$_1$, and third on Accuracy. 

The third system is RDI \cite{SemEval2016:task3:RDI} with MAP of 43.80, which is ranked third also on AvgRec and MRR. The system combines a TF.IDF module with a recurrent language model and information from Wikipedia.

\begin{table*}[tbh]
\small
\begin{center}
\begin{tabular}{@{}l@{ }@{ }l@{}}
\bf Team ID & \bf Team Affiliation\\
\hline
ConvKN       & Qatar Computing Research Institute, HBKU, Qatar; University of Trento, Italy\\
& \cite{SemEval2016:task3:ConvKN}\\
ECNU       & East China Normal University, China \\
& \cite{SemEval2016:task3:ECNU}\\
ICL00      & Institute of Computational Lingustics, Peking University, China\\
& \cite{SemEval2016:task3:ICL00}\\
ICRC-HIT   & Intelligence Computing Research Center, Harbin Institute of Technology, China\\
ITNLP-AiKF & Intelligence Technology and Natural Lang. Processing Lab., Harbin Institute of Technology, China\\
& \cite{SemEval2016:task3:ITNLP-AiKF}\\
Kelp       & University of Roma, Tor Vergata, Italy; Qatar Computing Research Institute, HBKU, Qatar\\
& \cite{SemEval2016:task3:KeLP}\\
MTE-NN     & Qatar Computing Research Institute, HBKU, Qatar\\
& \cite{SemEval2016:task3:MTE-NN}\\
overfitting & University of Waterloo, Canada\\
& \cite{SemEval2016:task3:overfitting}\\
PMI-cool   & Sofia University, Bulgaria\\
& \cite{SemEval2016:task3:PMI-cool}\\
QAIIIT     & IIIT Hyderabad, India\\
QU-IR      & Qatar University, Qatar\\
& \cite{SemEval2016:task3:QU-IR}\\
RDI\_team        & RDI Egypt, Cairo University, Egypt\\
& \cite{SemEval2016:task3:RDI}\\
SemanticZ  & Sofia University, Bulgaria\\
 & \cite{SemEval2016:task3:SemanticZ}\\
SLS        & MIT Computer Science and Artificial Intelligence Lab, USA\\
& \cite{SemEval2016:task3:SLS}\\
SUper\_team     & Sofia University, Bulgaria; Qatar Computing Research Institute, HBKU, Qatar\\
& \cite{SemEval2016:task3:SUper}\\
UH-PRHLT   & Pattern Recognition and Human Language Technologies Research Center,\\
           & Universitat Polit\`{e}cnica de Val\`{e}ncia; University of Houston\\
& \cite{SemEval2016:task3:UH-PRHLT}\\
UniMelb    & The University of Melbourne, Australia\\
& \cite{SemEval2016:task3:UniMelb}\\
UPC\_USMBA  & Universitat Polit\`{e}cnica de Catalunya, Spain; Sidi Mohamed Ben Abdellah University, Morocco \\
& \cite{SemEval2016:task3:UPCUSMBA}\\
\hline
\end{tabular}
\caption{The participating teams and their affiliations.}
\label{table:teams}
\end{center}
\end{table*}

\section{Features and Techniques}
\label{sec:features}


The systems that participated in several subtasks typically re-used some features for all subtasks, whenever possible and suitable.
Such features include the following: (\emph{i})~\emph{similarity features} between questions and comments from their threads or between original questions and related questions, e.g., cosine similarity applied to lexical, syntactic and semantic representations or distributed representations, often derived using neural networks,
(\emph{ii})~\emph{content features}, which are special signals that can clearly indicate a bad answer, e.g., when a comment contains ``thanks'',
(\emph{iii})~\emph{thread level/meta features}, e.g., user ID, comment rank in the thread,
and
(\emph{iv})~\emph{automatically generated features} from syntactic structures using tree kernels.

Overall, most of the top positions are occupied by systems that used tree kernels, combined with similarity features.
Regarding the machine learning approaches used, most systems chose SVM classifiers (often these were ranking versions such as SVM-Rank), or different kinds of neural networks. Below we look in more detail in the features and the used learning methods.

\subsection{Feature Types}

Participants preferred different kinds of features for different subtasks:

\paragraph{Subtask A.} Similarities between question subject vs. comment, question body vs. comment, and question subject+body vs. comment.

\paragraph{Subtask B.} Similarities between the original and the related question at different levels: subject vs. subject, body vs. body, and subject+body vs. subject+body.

\paragraph{Subtask C.} The same from above, plus the similarities of the original question ‚subject, body, and full levels‚ with the comments from the thread of the related question.

The similarity scores to be used as features were computed in various ways, e.g., the majority of teams used dot product calculated over word $n$-grams ($n$=1,2,3), character 3-grams, or with TF-IDF weighting.
Or simply using word overlap, i.e., the number of common words between two texts, often normalized, e.g., by question/comment length. Or overlap in terms of nouns or named entities.

Several systems, e.g., UH-PRHLT, KeLP, SLS, SemanticZ, ECNU, used additional similarities based on distributed representations. For example, using the continuous skip-gram model of word2vec or Glove, trained on Google News, on the English Wikipedia, or on the unannotated Qatar Living dataset.

In particular, UH-PRHLT used word alignments and distributed representations to align the words of the question with the words of the comment.

On the alignment topic, it is worth mentioned that MTE-NN applied a model originally defined for machine translation evaluation \cite{guzman-EtAl:2015:ACL-IJCNLP}, e.g., based on features computed with BLEU, TER, NIST, and Meteor \cite{ACL2016:MTE-NN-cQA}.
Similarly, ECNU used Spearman, Pearson, and Kendall Ranking Coefficients as similarity scores for question similarity estimation, whereas ICL00 used word-to-word translation probabilities, and UniMelb used convolutional neural networks (CNNs) fed with word embeddings and machine translation evaluation scores as input. 

ConvKN used a CNN that also encodes relational links between the involved pieces of texts
\cite{severyn2015sigir}. MTE-NN applied a simple neural network, ECNU and SLS used LSTM networks, and Overfitting applied Feedforward Neural Net Language Model (FNNLM).

It should be noted that ConvKN and KeLP used tree kernels with relational links \cite{Tymoshenko:2015:AIS:2806416.2806490,Tymoshenko:NAACL:2016}, i.e., the questions are aligned with the comments (or with the other questions) by means of a special REL tag, directly annotated in the parse trees.

Regarding text structures, UH-PRHLT used Knowledge Graph Analysis, which consists in labeling, weighting, and expanding concepts in the text using a directed graph. They also used frames from FrameNet to generate semantic features.

Several teams, e.g., ConvKN, KeLP and SUper Team, used meta-features, such as the user ID.
In particular, the SUper Team collected statistics about the users, e.g., the comments/questions they produced, time since their last activity, the number of good and bad comments in the training data, etc.

Other important features, which were used by most systems, are related to rank, e.g., rank of the comment in the question thread, or rank of the related question in the list of questions retrieved by the search engine for the original question.

Some exotic features by the SUper Team modeled readability, credibility, sentiment analysis, 
trollness \cite{mihaylov-georgiev-nakov:2015:CoNLL,mihaylov-EtAl:2015:RANLP2015,ACL2016:trolls}, and goodness polarity, e.g., based on 
PMI lexicons as for PMI-Cool.

Regarding Arabic, QU-IR and SLS used word2vec, whereas RDI\_Team relied on language models. 
In particular, the winning SLS team used simple text- and vector-based features, where the text similarities are computed at the word- and the sentence-level.
Most importantly, they computed two sets of features: one between the original and the related questions, and one between the original question and the related answer, which are then concatenated in one feature vector.

The ConvKN team combined some basic SLS features with tree kernels applied to syntactic trees, obtaining a result that is very close to that of the winning SLS team.

QU-IR used a standard Average Word Embedding and also a new method, Covariance Word Embedding, which computes a covariance matrix between each pair of dimensions of the embedding, thus considering vector components as random variables. 

Finally, RDI used Arabic-Wikipedia to boost the weights of medical terms, which improved their ranking function.

\subsection{Learning Methods}

The most popular machine learning approach was to use Support Vector Machines (SVM) on the features described in the previous section.
SVMs were used in three different learning tasks: classification, regression, and ranking.
Note that SVMs allow the use of complex convolutional kernels such as tree kernels, which were used by two systems (which in fact combined kernels with other features).

Neural networks were also widely used, e.g., in word2vec to train word embeddings.
As previously mentioned, there were also systems using CNNs, LSTMs and FNNLM.
Overfitting also used Random Forests.

Comparing tree kernels vs.~neural networks: approaches based on the former were ranked first and second in Subtask A, second and third in Subtask B, and second in Subtasks C and D, while neural network-based systems did not win any subtask, but neural networks contributed to the best systems in all subtasks, e.g., with word2vec.
Yet, post-competition improvements have shown that NN-based systems can perform on par with the best \cite{ACL2016:MTE-NN-cQA}.




\section{Conclusion}
\label{sec:conclusion}
We have described SemEval-2016 Task 3 on Community Question Answering, which extended SemEval-2015 Task 3 \cite{nakov-EtAl:2015:SemEval} with new subtasks (Question--Question similarity, Question--External Comment Similarity, and Reranking the correct answers for a new question), new evaluation metrics (based on ranking), new datasets, and new domains (biomedical for Arabic). The overall focus was on answering new questions that were not already answered in the target community forum.

The task attracted 18 teams, which submitted 95 runs; this is good growth compared to 2015, when 13 teams submitted 61 runs. The participants built on the lessons learned from the 2015 edition of the task, and further experimented with new features and learning frameworks. It was interesting to see that the top systems used both word embeddings trained using neural networks and syntactic kernels, which shows the importance of both distributed representations and linguistic analysis. It was also nice to see some new features being tried.

Apart from the new lessons learned from this year's edition, we believe that the task has another important contribution: the datasets we have created as part of the task (with over 7,000 questions and over 57,000 annotated comments), and which we have released for use to the research community, should be useful for follow up research beyond \hbox{SemEval}. 

Finally, given the growth in the interest for the task, we plan a rerun at SemEval-2017 with data from a new domain.

\section*{Acknowledgements} 
This research was performed by the Arabic Language Technologies (ALT) group at the Qatar Computing Research Institute (QCRI), HBKU, part of Qatar Foundation. It is part of the Interactive sYstems for Answer Search (Iyas) project, which is developed in collaboration with MIT-CSAIL.

We would like to thank the anonymous reviewers for their constructive comments, which have helped us improve the paper.

\bibliographystyle{naaclhlt2016}
\bibliography{naaclhlt2016}

\end{document}